\documentclass[twocolumn,prl,twoside]{revtex4-1}

\usepackage{amsmath, amssymb}
\usepackage{bm}
\usepackage{natbib}
\usepackage{color}

\usepackage{graphicx}
\usepackage{float}

\usepackage{todonotes}
\let\xtodo\todo
\renewcommand{\todo}[1]{\xtodo[inline,color=green!5]{#1}}

\begin{document}
\title{Neural network identification of people hidden from view with a single-pixel, single-photon detector}


\author{Piergiorgio Caramazza$^{1}$, Alessandro Boccolini$^{1}$, Daniel Buschek$^2$, Matthias Hullin$^3$, Catherine Higham$^4$, Robert Henderson$^5$, Roderick Murray-Smith$^4$ and Daniele Faccio$^{1}$}
\email{D.Faccio@hw.ac.uk}

\affiliation{$^1$Institute of Photonics and Quantum Sciences, Heriot-Watt University, Edinburgh EH14 4AS, UK}
\affiliation{$^2$Media Informatics Group, University of Munich (LMU), Munich, Germany}
\affiliation{$^3$Institute for Computer Science II, University of Bonn, Friedrich-Ebert-Allee 144 53113 Bonn, Germany}
\affiliation{$^4$Institute for Micro and Nano Systems, University of Edinburgh , EH8 9YL, Edinburgh, United Kingdom}
\affiliation{$^5$School of Computing Science, University of Glasgow, Glasgow G12 8QQ, UK}

\begin{abstract}
{{Light scattered from multiple surfaces can be used to retrieve information of hidden environments. However, full three-dimensional retrieval of an object hidden from view by a wall has only been achieved with scanning systems and requires intensive computational processing of the retrieved data.  Here we use a non-scanning, single-photon single-pixel detector in combination with an artificial neural network: this allows us to locate the position and to also simultaneously provide the actual identity of a hidden person, chosen from a database of people {($N$=3)}.  Artificial neural networks applied to specific computational imaging problems can therefore enable novel imaging capabilities with hugely simplified  hardware and processing times.}}


\end{abstract}

\maketitle

\begin{figure}[t!]
\centering
\includegraphics[width = 8cm]{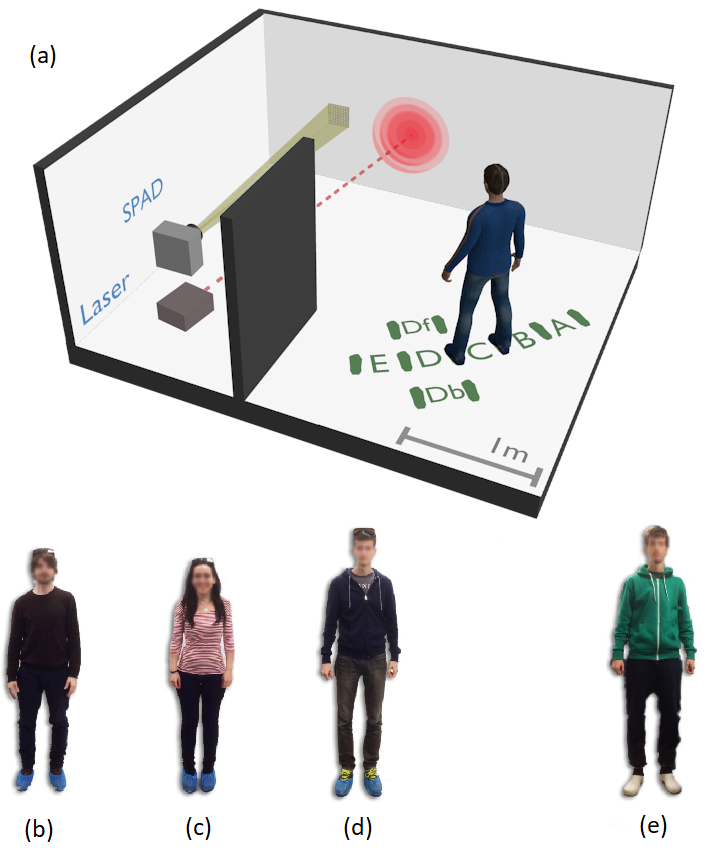}
\caption{Experimental details: layout is shown in (a). A pulsed laser light source illuminates a wall which scatters light behind the obscuring wall, thus illuminating the hidden person. The return echoes are collected by the SPAD array, aimed at the first scattering wall.  The three people used for the experiments, referred to in the text and figures as ``n.1'', ``n.2'' and ``n.3'' are shown in (b), (c), (d), respectively.   Two different case were verified: different clothing, (b)-(d), and same clothing (e) (only individual ``n.3'' is shown for simplicity). Measurements where repeated 5 times across all 7 different positions.
\label{fig:layout}}
\end{figure}

\begin{figure}[t]
\includegraphics[width = 8cm]{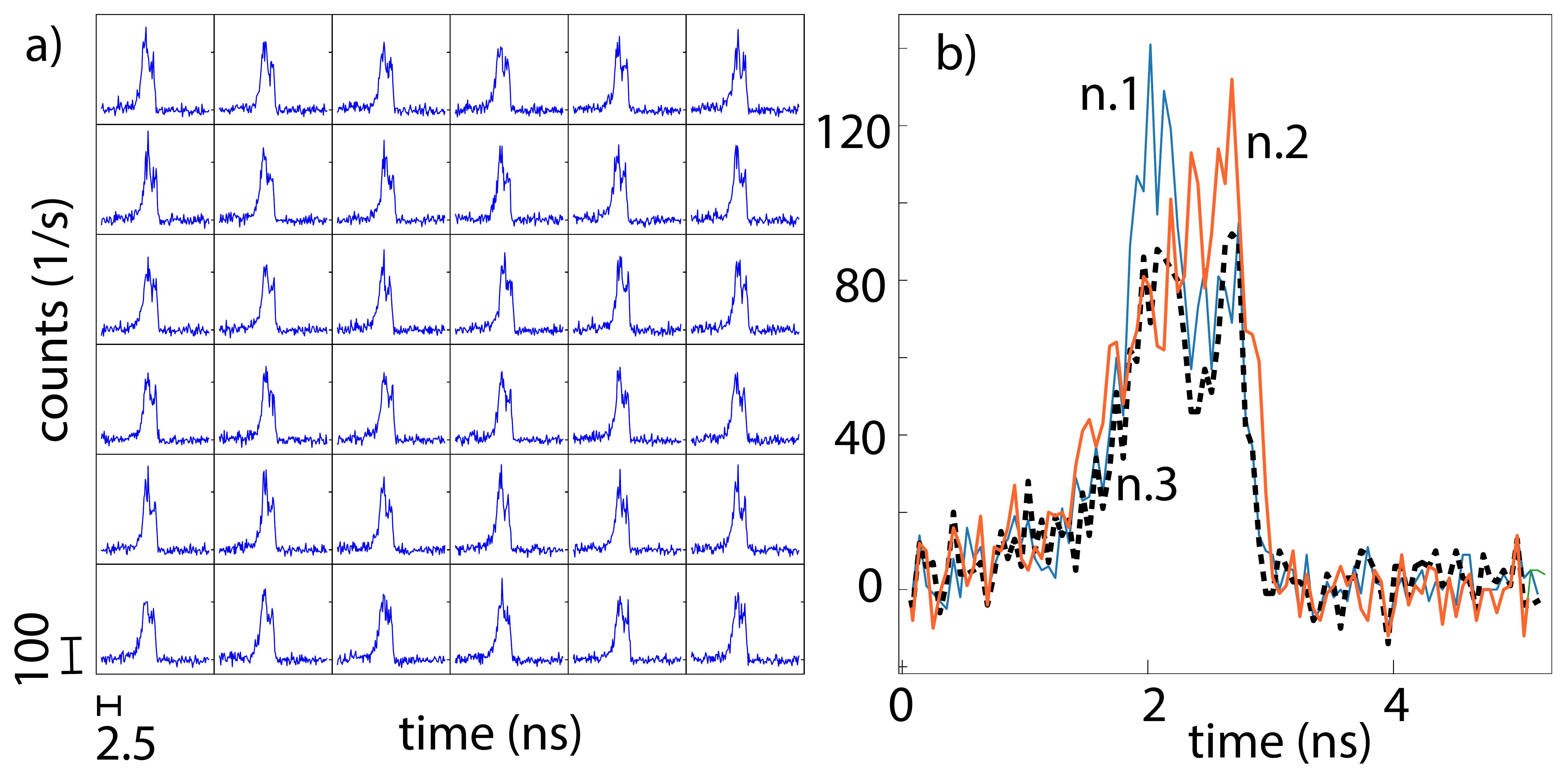}
\caption{(a) Example of the input data for the ANN, showing a subset 6x6 taken from the full 32x32 array. This data is a single measurement of individual ``n.1'' in position ``C''. (b) For comparison, we show the temporal histograms for individuals n.1, 2 and 3 for the same pixel.
\label{fig:histograms}}
\end{figure}

{\bf{Introduction.}}
Recent years have seen a surge of interest and corresponding advances in the ability to image objects that are not visible within the direct line of sight. In particular we refer to  the situation in which the object is hidden behind a wall, a corner or inside a room to which we do not have access \cite{2K2009,6HSO2012,velten,me2,buttafava2015non,klein2016tracking,katz2012looking,10TOFSMIT2016,chan}. To date, all techniques that attempt to image a scene or object that is hidden behind an obstacle have relied on active imaging, i.e. the scene is actively {illuminated} using a light source that is controlled by the observer. Although recent work used continuous illumination \cite{klein2016tracking}, the most common approach is to use a pulsed light source, for example a laser. The basic functioning principle is then very similar to listening to sound echoes reflected from multiple surfaces: the laser beam is scattered off a surface that lies within the direct line of sight, but also such that the scatter may enter the hidden environment. By then synchronising the detection system to the emitted pulses and measuring the return time for each echo, it is possible to determine the distance of the object that created/reflected the signal. If one wants to build a full image of the hidden environment or object, simply measuring return times from a single point is not sufficient: multiple pixel information is required and is built up by either directly imaging and/or scanning the imaging optics across the surface where the reflected echoes are detected, or by scanning the illumination spot on the first scattering surface. Both approaches, followed by computational processing of the collected data can provide full 3D reconstruction of the hidden environment.\\
However, the overall constraints on the problem make it extremely hard to achieve 3D imaging of hidden objects with high resolution (few mm or less), at significant distances (1 m or more) and within reasonable (e.g. less than several seconds) time-frames. The main limitations are: the very low return signal which will typically decay as $\sim1/d^6$ ($d$ is the distance of the hidden object from the imaging system) \cite{chan};  the very high temporal resolution (10-100 ps or less) required for the detector to obtain sub-cm precision; the requirement to scan either the laser or the detector lengthens the acquisition times to minutes or even hours \cite{buttafava2015non,velten}. Currently, there is no obvious and single solution to all of these problems: 3D imaging can be obtained at the expense of acquisition and processing time or tracking can be obtained at higher frame rates, albeit with limited resolution and hence the impossibility to reconstruct actual 3D shapes of the hidden objects. Moreover, very little work has been performed to date on large objects and actual people, with results limited to tracking of location and movement detection \cite{chan}.\\
If we then ask the question, ``is it possible to both locate and even identify a person {who} is hidden behind a wall?'', we are faced with even larger hurdles and the answer is that such a capability is well out of the reach of all currently adopted approaches. \\
The key point of this work is to show that by moving beyond the current paradigms of non-line-of-sight (NLOS) imaging, it is actually possible to perform the feat questioned above on at least a limited set of chosen targets. In order to do this we use data from a single-photon, single-pixel detector that is analysed by a previously trained artificial neural network (ANN). 
Use of just one single pixel detector is highly advantageous due to commercial availability and optimised specifications in terms of photon sensitivity (can be close to 100\%) and temporal resolution (can be lower than 20 ps). The ANN is able to recognise the position of a hidden person (from a training set of 7 positions) and even provide the identity of the person (from a training set of 3 people). Remarkably, person-identification is successful even when all three individuals have the same clothing, thus hinting that the ANN can recognise the more subtle changes in the physiognomy from one person to another. This is all the more remarkable when we note that the temporal resolution of the detector (120 ps, corresponding to 1.8 cm depth resolution) would not be sufficient to precisely reconstruct the full 3D shape of a face, even after raster scanning. The ANN therefore extends the capability of the imaging system and by removing the requirement of any forward modelling or computational post-processing, allows fast location and identification of hidden people that would otherwise not be possible.\\
\begin{figure}[t!]
\centering
\includegraphics[width = 5cm]{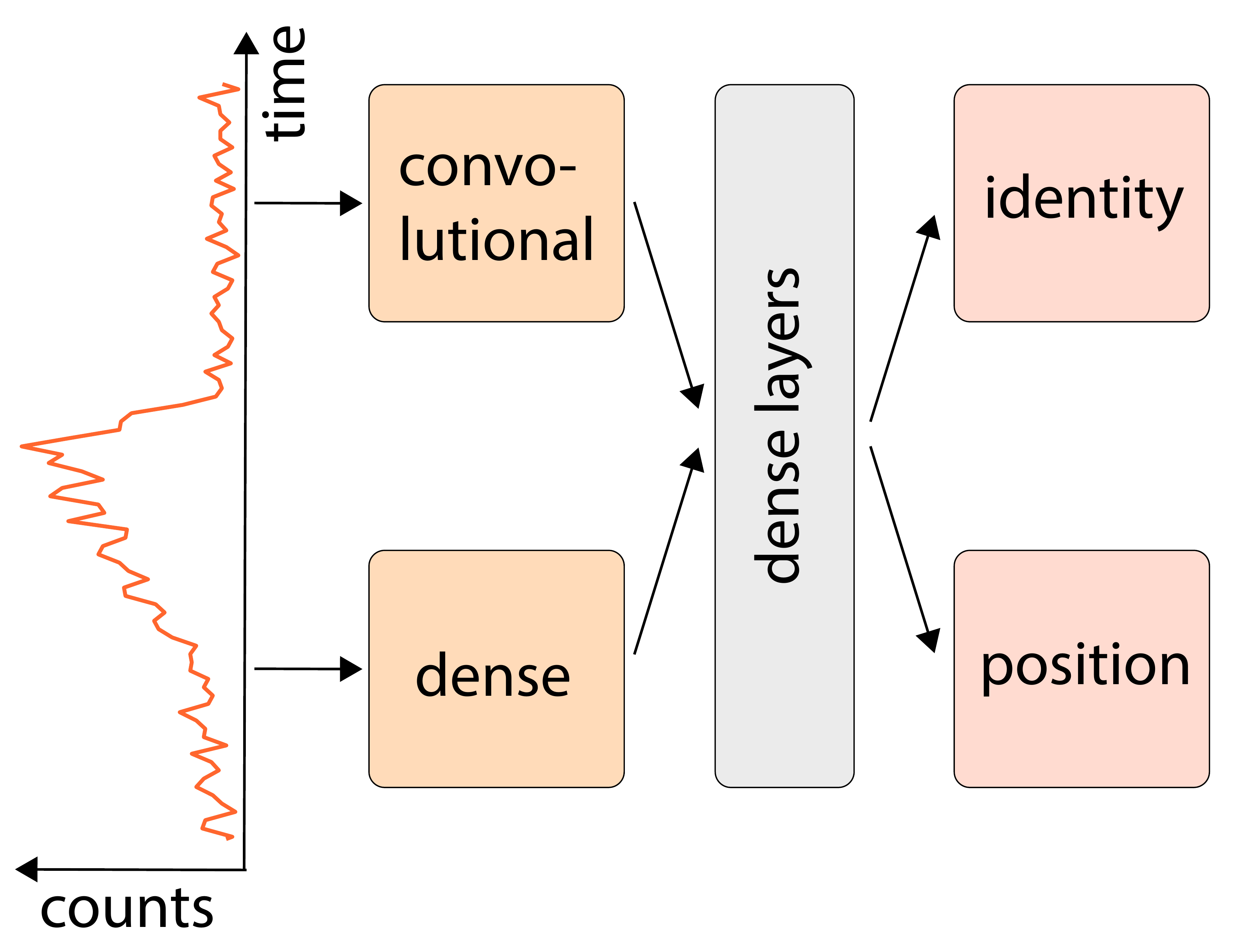}
\caption{Network architecture tested and discussed in the text (1Dconv +  Dense).
\label{fig:networkarch}}
\end{figure}
{\bf{Experiments.}}
Data is collected with the set-up shown in Fig.~\ref{fig:layout}. A pulsed laser beam, $\lambda = 808$ nm and Rep. Rate $=80$ MHz, is pointed towards a wall where a first scattering occurs. Thus, the light illuminates the hidden individual so that the backscattered light can be captured by the detection system. Three different people have been used for this experiment, and both same clothing and different clothing data were acquired (see Fig.~\ref{fig:layout}). Moreover, we tested seven different positions of which, those from ``A" to ``E" share similar photon time-of-flight (i.e. the individual has a similar distance from the first scattering point on the wall) This situation was chosen so as to ensure that the ANN did not train solely on arrival time of the return photon echoes but rather focused on actual features within the temporal shape of the echo. For each person in each position, five separate measurements were taken, alternating people and position for each measurement.\\
In order to provide the large amount of data required for the neural network training, we make use of a single photon avalanche diode (SPAD) segmented array \cite{spad}. This consists of a 32x32 array of single-pixel SPAD detectors, each characterized by an instrumental temporal response function (IRF) of $\sim 120$ ps. Thus, the pixels are treated as independent observers that are looking at roughly the same position on the wall (within the 3x3 cm$^2$ imaged area on the wall). We also acquire a background signal under identical conditions but with the individual removed: this is then subtracted from the measurements. Furthermore, we eliminate ``hot'' pixels and thus obtain $800$ temporal histograms from each measurement. A subset example of data from a single, background-subtracted measurement (for individual ``n.1" in position ``C")  is shown in  Fig.~\ref{fig:histograms}a).  The SPAD array is triggered directly from the laser external trigger and is thus synchronised to the emission of each individual laser pulse: a single acquisition takes two seconds, equivalent to integration over $80\times10^6$ laser pulses.  \\
\noindent{\bf Analysis of experimental data}
The architecture we propose is inspired by the physics involved in the experiment. Here, the underlying assumption is that the information about position and shape of the hidden individuals are both encoded in the photon time-of-flight and final temporal shape of the return echo.  In Fig.~\ref{fig:histograms}b) we show a typical example of the histograms for the three individuals tested in this work (labelled as n.1, n.2 and n.3) As can be seen, there are clear differences between the three temporal histograms yet there is no unique feature that stands out as distinguishing one from the other. They have similar heights, total photon counts (physically connected to the overall target reflectivity) and widths (physically connected to the overall height of the target individual). This means that any data-driven classification approach has to learn to identify the overall ensemble of more subtle differences and accordingly classify the data accordingly. \\
\noindent{\bf{Neural Network classifier.}}
{Artificial neural networks (ANN) are mathematical models loosely inspired by the human brain, which have provided an important contribution to both scientific and technological research for their capacity to learn distinctive features from large amounts of data. Deep convolutional neural networks are computational models which are concerned with learning representations of data with multiple levels of abstraction. They have been broadly employed for tasks such as regression, classification, unsupervised learning, and are proving very successful at discovering features in high-dimensional data arising in many areas of science,  with breakthroughs in image processing and time-series analysis \cite{Sch15, DL_nature,GooBenCou16}.  The performance increases are  due to increases in processing power, algorithm improvements better quality flexibility software, and the availability of large collections of training data.} 

Very recent studies have started to look at the use of ANNs in the area of computational imaging with applications in phase-object identification \cite{mit}, pose-identification of human-shaped objects \cite{satat2017object} and number/letter identification \cite{Lyu} from  behind a diffusive screen. In this work, we rely on supervised machine learning algorithms in order to classify people hidden from our line of sight and located in varying positions.

{We therefore build a nonlinear classifier aiming to correctly identify the label of the histograms resulting from the acquisition of pulsed laser light backscattered from three different people in seven different positions.  We use a supervised approach where we pair the temporal histogram as input to the ANN and create an output vector encoding the class of the person and the target location. Both class and location are treated as categorical classification tasks, and encoded using a `one-hot' encoding such that we use $N_c$ binary outputs for $N_c$ classes, and $N_l$ binary outputs for location positions. In this work $N_c=3, N_l=7$. The cost function minimised during learning is the categorical cross-entropy \cite{Keras2015}.

The ANN architecture processes input data in parallel through: a fully-connected layer on one side, in order to retrieve more information about the distance, and in parallel, convolutional layers which due to their translation invariant nature, will focus more on the temporal histogram shape and features.\\

{\bf{Results.}} 
{After optimisation was completed, we tested the performance of the classifier on new data, taken under the same conditions as the training data but is not used during the training process. To test the robustness of the optimisation process, we use a leave-one-out cross-validation process, where we extract all data from one illumination to use as test data, then train on data associated with all other illuminations. We then average the classification results in a per-pixel basis. For this training set with 5 illuminations of ca. 800 pixels each, the classification average calculated over  5 runs obtained by training the ANN on measurements from 4 illuminations and testing on the remaining one (repeating the procedure by all permutations of the 4 training and single test data sets).}
\begin{figure}[t]
\centering
\includegraphics[width = 8cm]{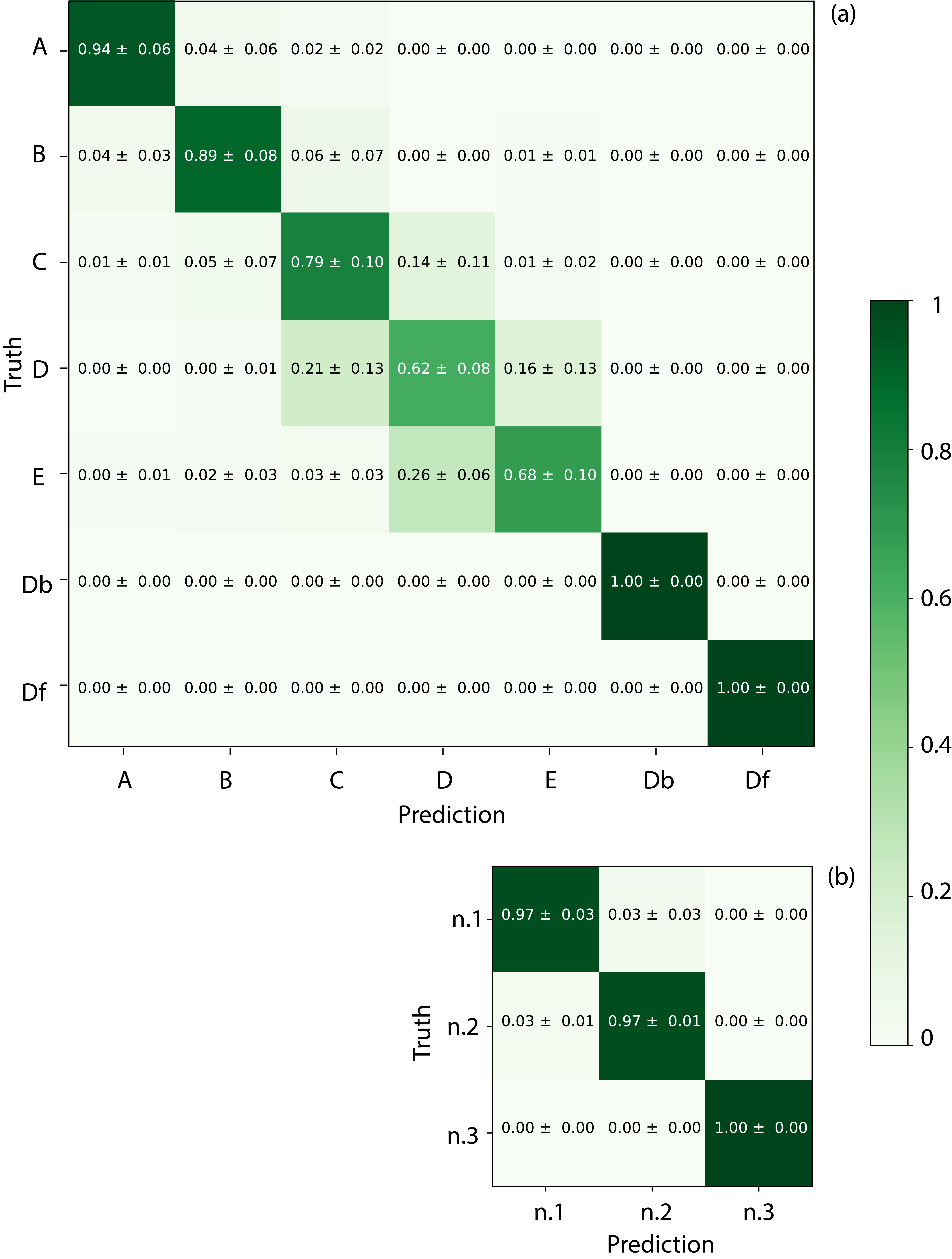}
\caption{Results for the case in which the three individuals (n.1, n.2, n.3) have {\emph{different}} clothing. Confusion matrices are shown for: (a) retrieval of position (averaged over all individuals) and (b) retrieval of the individual identities. The ANN was trained with 4 measurements and tested with one, repeating over every permutation and averaging.
\label{fig:different_clothes}}
\end{figure}
\begin{figure}[t]
\centering
\includegraphics[width = 8cm]{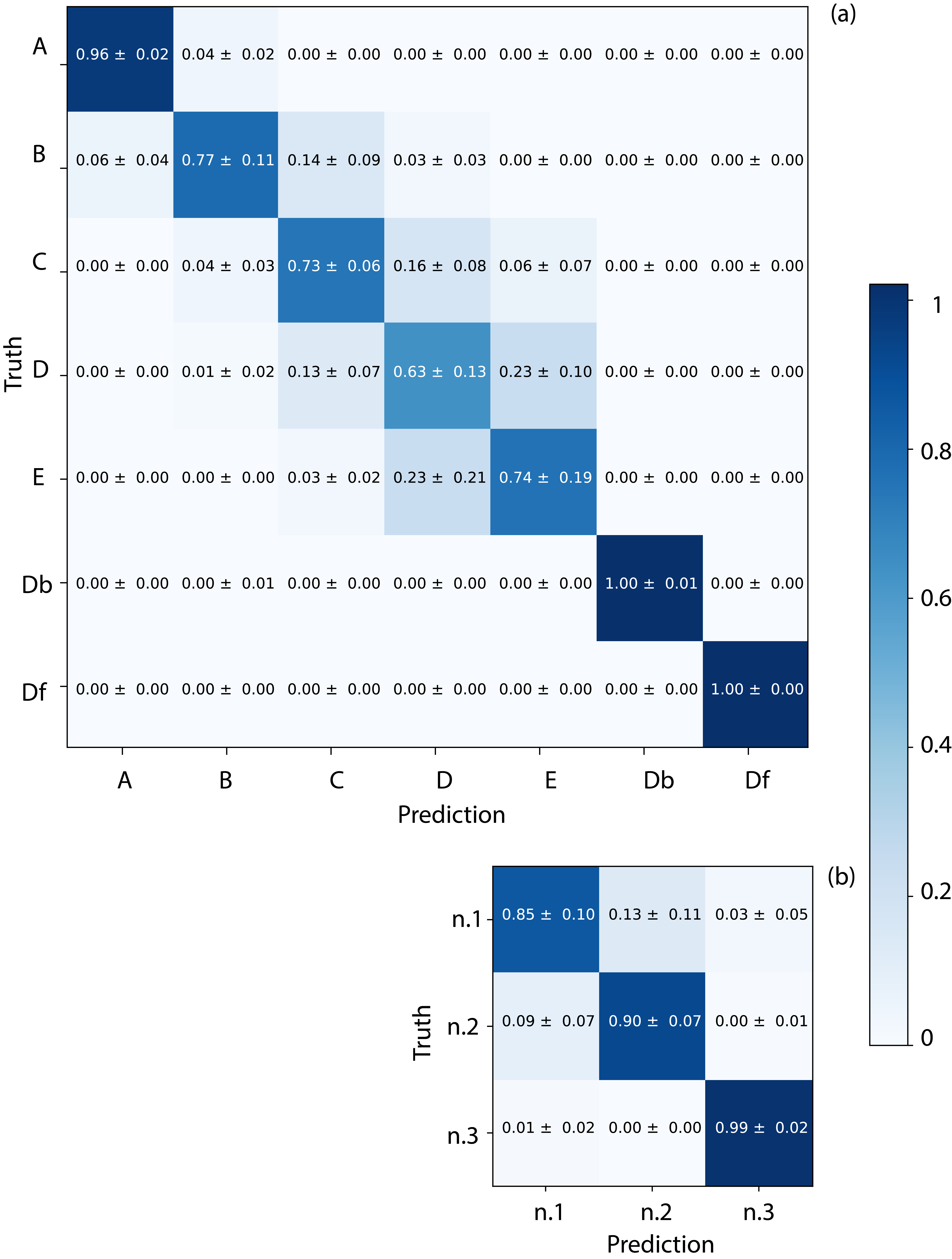}
\caption{Results for the case in which the three individuals (n.1, n.2, n.3) have the {\emph{same}} clothing. Confusion matrices are shown for: (a) retrieval of position averaged over results for all three individuals and (b) retrieval of the individual's identity.
\label{fig:same_clothes}}
\end{figure}
Firstly, we show the results for the case in with all three individuals have ``different clothing'', Fig.~\ref{fig:different_clothes}. Data are reported in a confusion matrix that compares the actual classes (vertical axis, ``truth'') with the predicted classes (horizontal axis). {The matrix values represent the normalised average calculated over all 5 cross-validation runs}. A ``0'' indicates no overlap between training and test data and a ``1'' indicates 100\% statistical certainty in the correct classification.\\
As we can observe in  Fig.~\ref{fig:different_clothes}(a), positions ``Db'' and ``Df'' are identified with 100\% certainty, whereas among the positions with similar photon time-of-flight, the classification fidelity is slightly worse, thus indicating a certain role played by the overall arrival time of the photons. However, the classification still shows a very good agreement for all positions with the ground truth. In  Fig.~\ref{fig:different_clothes}(b) we show the confusion matrix for person identification indicating that the classifier is able to correctly identify the three people. Since it is possible in this case that both shape and reflectivity of the bodies (that are clothed differently) play a key role, we try to isolate one of these two degrees of freedom by repeating the experiments with all individuals in the same clothing, see Fig.~\ref{fig:same_clothes}. 
The classification of the individual's position occurs with similar fidelity as in the ``different clothing'' case, Fig.~\ref{fig:same_clothes}(a). However, person identification is indeed now more challenging with a higher confusion between individuals. The notable result from these tests is that even when individuals have the same clothing, a single-pixel is sufficient to identify them. All of the information is encoded in the temporal shape of the return photon echo. {Yet, as the temporal resolution of 120 ps corresponds to a spatial depth resolution of 1.8 cm which is insufficient to create a 3D map of a human face, so we suspect that the classifier is not focusing solely on facial features, but is relying on overall physiognomy, including body height, width and possibly also skin reflectivity.}\\

Finally, we compared the results from different ANN architectures. The results 
 tend to not show any particular sensitivity to the specific ANN architecture employed, therefore suggesting the robustness of this approach and therefore that further improvement would need to come from larger or more controlled training sets. However, the ANN performance  does seem to suggest that classifying location and identities jointly is consistently better than trying to deal with these individually. {This is probably because the internal representations learned to predict class can then be useful to help predict location more accurately, and vice versa.}

Instead of taking average classification performance at a per-pixel level, we can base classification on a majority verdict, all ca 800 per-pixel classifications for a single illumination. The results of this approach 
indicate that misclassification occurs within certain illuminations rather than across illuminations suggesting that increasing the variation in the training data should improve the classifier.\\
{\bf{Conclusions.}}
\noindent One-dimensional temporal histograms obtained by capturing laser echoes backscattered from a hidden body contain information that reliably allow the identification of different people in different positions. We underline that once the ANN is trained, the classification process is just the result of multiple matrix
multiplications {and vector function evaluations} and can thus proceed extremely quickly, with millisecond processing times. {Thus, a data-driven classifier, such as the ANN used here,} can achieve identification with a precision that is not possible with any other currently available approach, and it does so with speeds that are orders of magnitude faster than even the best NLOS reconstruction shown to date. 

Interesting questions arise from this work, such as exactly how many individuals may be used for training and then successfully identified with a single temporal histogram that has a given IRF, and the impact of target movement on the classification performance. The information content analysis of temporal photon echoes remains an open question for future work. Recent work has also shown how device-independent training is possible within the context of computational imaging problems, promising the ability to either train and test using completely different detectors or even to train using forward modelling as in \cite{satat2017object}. The latter would be particularly significant in the context of classification of hidden environments where the forward model is possibly easier to fully characterise and model with respect to people. These results pave the way to exciting novel scenarios for machine learning applications, such as identification of groups of individuals (for example distinguishing adults from children) but also of entire environments by means of a single pixel temporal measurements.


\end{document}